\documentclass[10pt,journal]{IEEEtran}
\usepackage{textcomp}
\usepackage[pdftex]{graphicx}
\graphicspath{{../}}
\DeclareGraphicsExtensions{.pdf,.jpeg,.png,.eps}
\usepackage{epstopdf}
\usepackage{algorithm} 
\usepackage{algpseudocode} 
\usepackage{amsmath}
\usepackage[frenchb, english]{babel}
\usepackage{booktabs}
\usepackage{cite}
\usepackage{caption}
\usepackage{subcaption}
\usepackage{float,multirow}
\usepackage{amsfonts}
\usepackage[none]{hyphenat}
\usepackage{url}

\usepackage{breakurl}
\usepackage[breaklinks]{hyperref}
\usepackage{soul}
\usepackage{dirtytalk}
\usepackage{hyperref}
\usepackage{comment}
\usepackage[euler]{textgreek}
\usepackage{epsfig,bm,amsmath,amssymb,graphicx,setspace}
\usepackage{algorithm}
\usepackage{color}
\usepackage[T1]{fontenc}
\usepackage{algpseudocode}
\usepackage{textgreek}
\usepackage{amsthm}
\usepackage{array}
\usepackage{balance}
\usepackage{enumerate}
\usepackage[utf8]{inputenc}
\usepackage{array}
\usepackage{wrapfig}
\usepackage{multirow}
\usepackage{tabularx}

\usepackage[section]{placeins}
\usepackage{wasysym}
\usepackage{graphicx}
\usepackage{subcaption}
\usepackage[utf8]{inputenc}
\usepackage[export]{adjustbox}
\usepackage{wrapfig}

\usepackage{array}
\usepackage{multirow}
\usepackage{graphicx}
\usepackage[table,xcdraw]{xcolor}
\usepackage{amssymb}

\def\BibTeX{{\rm B\kern-.05em{\sc i\kern-.025em b}\kern-.08em
        T\kern-.1667em\lower.7ex\hbox{E}\kern-.125emX}}
\usepackage[T1]{fontenc}
\usepackage{multirow}
\usepackage[frenchb]{babel}
\newcolumntype{C}[1]{>{\centering\arraybackslash}m{#1}}
\newcolumntype{P}[1]{>{\centering\arraybackslash}p{#1}}

\begin{document}

  \title{A Novel Cloud-Based Diffusion-Guided Hybrid Model for High-Accuracy Accident Detection in Intelligent Transportation Systems}
    
  \author{Siva Sai*, Saksham Gupta*,  Vinay Chamola \textit{Senior Member, IEEE}, Rajkumar Buyya \textit{Fellow, IEEE}
\thanks{* Equal contribution }
\thanks{This work was supported by CHANAKYA Fellowship Program of TIH Foundation for IoT \& IoE (TIH-IoT) received by Dr. Vinay Chamola under Project Grant File CFP/2022/027. }

 \thanks{Siva Sai, Saksham Gupta and Vinay Chamola are with the Department of Electrical and Electronics Engineering, BITS-Pilani, Pilani Campus, 333031, India. (e-mails: \{p20220063, f20220758, vinay.chamola\}@pilani.bits-pilani.ac.in). Vinay Chamola is also with APPCAIR.}
 \thanks{Rajkumar Buyya is with The Quantum Cloud Computing and Distributed Systems (qCLOUDS) Laboratory, School of Computing and Information Systems, The University of Melbourne, Australia (e-mail: rbuyya@unimelb.edu.au). Siva Sai is also with qCLOUDS Laboratory.}
 }

  % \thanks{Digital Object Identifier: XXXXXXXXXXXX}
  % }

\maketitle

\begin{abstract}    
The integration of Diffusion Models into Intelligent Transportation Systems (ITS) is a substantial improvement in the detection of accidents.
We present a novel hybrid model integrating guidance classification with diffusion techniques. By leveraging fine-tuned ExceptionNet architecture outputs as input for our proposed diffusion model and processing image tensors as our conditioning, our approach creates a robust classification framework.
Our model consists of multiple conditional modules, which aim to modulate the linear projection of inputs using time embeddings and image covariate embeddings, allowing the network to adapt its behavior dynamically throughout the diffusion process.
To address the computationally intensive nature of diffusion models, our implementation is cloud-based, enabling scalable and efficient processing.
Our strategy overcomes the shortcomings of conventional classification approaches by leveraging diffusion models inherent capacity to effectively understand complicated data distributions. We investigate important diffusion characteristics, such as timestep schedulers, timestep encoding techniques, timestep count, and architectural design changes, using a thorough ablation study, and have conducted a comprehensive evaluation of the proposed model against the baseline models on a publicly available dataset. The proposed diffusion model performs best in image-based accident detection with an accuracy of 97.32\%.
\end{abstract}  
% Later => Add results in abstract
% We have also conducted several ablation studies experimenting with the number of clients and the cut layer position in the split learning framework.
% Although federated learning solves data privacy and related issues, it has disadvantages
\begin{IEEEkeywords}
  Diffusion Models, Accident Detection, Intelligent Transportation Systems, LLMs.
\end{IEEEkeywords}

\section{Introduction}
Intelligent Transport Systems (ITS) have benefited greatly from the rapid growth of artificial intelligence (AI), which has had a huge impact on many other industries \cite{advanced_learning_technologies_ITS}. Large Language Models (LLMs) have become revolutionary tools in this field, providing previously unheard-of processing and interpretation capabilities for complicated data \cite{llm_intelligent_transportation_review}. Applications ranging from traffic control to autonomous driving have been transformed by LLMs, which are distinguished by their capacity to comprehend and produce language that is human-like \cite{llm_intelligent_transportation_review, embracing_llms_traffic_flow_forecasting}. Their incorporation with ITS offers better safety protocols, more effective transportation networks, and improved decision-making procedures \cite{advanced_learning_technologies_ITS}.

Because of their adaptability, LLMs may be used in a variety of ITS contexts. To improve traffic flow and lessen congestion, LLMs may evaluate enormous volumes of unstructured data from sources like GPS, traffic sensors, and security cameras \cite{towards_explainable_traffic_flow}. In order to improve the navigation skills of driverless cars, organizations like Waymo are investigating the use of multimodal LLMs, like Google's Gemini, to interpret sensor data and predict future trajectories \cite{multimodal_llm_intelligent_transportation}. Additionally, by facilitating better human-machine interactions, LLMs let drivers and in-vehicle technologies communicate more naturally.

Accident detection is a crucial ITS domain where LLMs may make a significant contribution. Reducing reaction times, minimizing traffic interruptions, and improving overall road safety all depend on accurate and rapid accident detection \cite{vision_based_traffic_accident_detection}. Conventional accident detection techniques have depended on preset algorithms and organized data, which could not adequately represent the complexity of real-world situations \cite{advanced_learning_technologies_ITS}. In order to better detect and evaluate accident scenarios, the integration of LLMs offers a fresh method by utilizing their capacity to analyze and understand unstructured data, such as textual accident reports and social media feeds \cite{llm_intelligent_transportation_review}.

A key component of ITS is accident detection, which is essential to maintaining traffic safety and streamlining traffic control. Rapid emergency responses, fewer secondary events, and less traffic congestion are all made possible by early and precise accident detection \cite{vision_based_traffic_accident_detection}. Traffic cameras, sensors installed on roads, and manual reporting systems have all been used in traditional accident detection methods \cite{advanced_learning_technologies_ITS}. Although somewhat successful, these approaches frequently have drawbacks in terms of coverage, real-time processing, and the capacity to decipher intricate situations \cite{vision_based_traffic_accident_detection}.
AI-driven techniques have recently been included to improve accident detection skills. For example, machine learning algorithms may evaluate sensor data to discover abnormalities suggestive of collisions, and computer vision techniques can be used to analyze video feeds to identify accidents \cite{advanced_learning_technologies_ITS}. The dynamic and unstructured nature of real-world data, however, may be difficult for these models to handle, and they frequently require intensive feature engineering. Furthermore, difficulties in incorporating sophisticated AI models into real-world applications have been noted, including high computational costs and constrained processor power \cite{vision_based_traffic_accident_detection}.

There are still a number of issues with accident detection models, even with the improvements made by conventional machine learning approaches. These include issues with managing unstructured data, processing in real-time, and adjusting to various and changing traffic situations \cite{advanced_learning_technologies_ITS}. Diffusion models have been suggested as a viable way to address these issues \cite{leveraging_llms_urban_intersections}. 
% Updated:
Our objective is to tackle classification problems through a generative modeling framework utilizing diffusion models, leveraging the inherent stochasticity of its outputs to improve the accuracy of an underlying deterministic classifier.    
In comparison to other generative architectures like GANs and VAEs, diffusion models capture a richer latent space representation owing to the introduction of timesteps and UNET architectures.
GANs although generate high-fidelity samples, but do not match the distribution of real data as effectively, whereas VAEs, while good for representation learning and uncertainty estimation, struggle to generate samples as sharp and detailed as diffusion models.
Hence, the use case of classification via generative architectures is best suited to Diffusion models.
Diffusion models can be integrated into ITS to improve planning and decision-making, which will result in more precise and faster accident detection.

In order to provide a comprehensive grasp of traffic conditions and possible accident situations, diffusion models may also be expanded to incorporate data from a variety of sources, including social media, meteorological information, and traffic sensor data. In summary, ITS's use of diffusion models and LLMs marks a substantial improvement in accident detection techniques. By using these models' advantages, ITS can identify accidents more thoroughly, accurately, and quickly, which will eventually improve road safety and make transportation systems more effective.
Our proposed model combines a guidance classifier with a diffusion classification approach. It utilizes softmax outputs from a fine-tuned ExceptionNet model as the mean for the diffusion model's final timestep, while image tensors serve as covariates processed through an encoder. The architecture features multiple Conditional Module layers incorporating timestep embeddings, enhanced with softmax activation functions and batch normalization. A key checkpoint multiplies the image encoder's output features with processed labels before proceeding through four Conditional Module blocks to produce the final output.
Our proposed model showed better Accuracy, F1-Score, Precision, and Recall values compared to conventional CNN architectures, namely ResNet-50, ResNet-18, and GoogLeNet. Our model showed the highest accuracy of 97.32\%.

The main contributions of the paper are listed as follows: 
\begin{itemize}
\item We introduce a novel approach to vehicle accident detection using a diffusion classification model.
\item We explore various diffusion parameters in our ablation study, such as the choice of timestep scheduler, how we encode the timestep information, and the diffusion architecture we used.
\item We experiment with the number of timesteps used and their effect on the training and inference.
\end{itemize}

The rest of the paper is organized as follows. In section \ref{relworks}, we cover the works related to the proposed model. We discuss the proposed methodology in Section \ref{method} followed by a discussion on the experimentation results in Section \ref{results}. Finally, we conclude the paper in Section \ref{concl}.

\section{Related Work}\label{relworks}
% In this section, we discuss existing works related to our proposed model.
% \subsection{Classification Models in ITS}
% 

The application of classification models has received tremendous domain application in the field of accident detection. Earlier work focused on traditional machine learning pipelines like Random Forest, SVM, and gradient boosting \cite{deep_neural_framework_accident_detection} \cite{ml_framework_automated_accident_detection}. With the advent of high parameter and stronger models like deep learning such as MLP and robust parametric efficient models like CNNs, more works followed.
Paper \cite{small_parallel_residual_cnn_traffic_congestion} proposes a CNN framework to recognize congestion in varied scenarios using visual data. The architecture of the proposed model include parallel residual convolutional units, inspired by the ResNet architecture.
Paper \cite{review_yolo_cnn_real_time_accident_detection} tackles real-time accident detection using a combination of YOLO \cite{review_yolo_cnn_real_time_accident_detection} and CNN frameworks which work together to improve scalability by designing the system to be compatible with multiple camera feeds for urban environments.
Other notable directions include the introduction of ViTs (Vision Transformers). Paper \cite{vision_transformer_multi_phase_accident_detection} introduces multi-phase frameworks that combine YOLO, ByteTrack, and Vision Transformers to precisely detect accident events and mitigate false positives in real-world traffic ideas, showing superiority over standard CNN-based models.

Due to high generalizability and scope of representation learning in diffusion models, Generative Adversarial Networks supersede previous models used to enhance classification tasks.
% \subsection{Generative Models in ITS}
% The field of traffic detection has greatly benefited and has achieved tremendous success in the application of generative models. Since the introduction of VAEs, generative models have been extensively utilized to the benefit. The methods introduced by (\cite{KingmaWelling2014AutoEncodingVB}) integrate a novel approach to unsupervised learning that efficiently optimizes the variational lower bound on marginal likelihood. It also combines deep learning with probabilistic modeling by learning a continuous latent representation of data using variational inference. Applications like (\cite{HuZhang2022VariationalAutoencoderGraphConv}) for robust medium/long-term prediction of traffic flow and speed from sensor data. The authors use multiple techniques including fusing of local spatial-temporal features with graph convolution to understand traffic node relationships over time learning process to maximize the variational evidence lower bound (ELBO) and minimize the KL divergence.
% Further, GANS (\cite{Arjovsky2017WGAN}) address limitations of traditional generative models by avoiding explicit density modeling and intractable likelihood computations that plagued earlier approaches. Unlike VAEs which used reconstruction loss, GANs used adversarial loss for more realistic sample generation.
Paper \cite{Zhang2023GANEnhancedTrafficAccidentDetection} leverages the genius of GANs for accident detection. Designed for real-time accident detection from CCTV, the approach is suited for smart city surveillance and emergency management integration. The authors tackle the problem of accident detection using three models - standard CNN, fine-tuned CNN (FTCNN), and Vision Transformer (ViT).
% Recent works use variation of GANs such as (\cite{Zhang2023GANEnhancedTrafficAccidentDetection}) where the authors propose an improved GAN model customized for non-image traffic accident feature data. The generator used in the framework is inspired by the DenseNet architecture due to multiple short connections. However, the method used is not image-based, limiting direct application to vision-based accident analysis. (\cite{Lai2020GANDayToNightVehicleDetection}), on the other hand, uses multi-sensory data and employs a CNN-based feature extraction from environment sensor data module and focuses on multimodal sensor fusion for anomaly detection.
% Multiple variations of GAN architecture have been popularized throughout the years including (\cite{Arjovsky2017WGAN}) where vanishing gradients from the original GAN formulation were addressed and theoretical justification for enhanced training stability was provided, and TGAN (\cite{Radford2016TGAN}) which handles sequences and temporal data, suitable for time-series data which is widely available in video-level accident detection analysis where each frame subsequently influences the other.
% For instance, this method addresses the problem of domain shift between day and night with the variation of GAN architectures. Instead of vanilla GANs, the authors (\cite{Lai2020GANDayToNightVehicleDetection}) utilize CycleGAN (\cite{Zhu2017CycleGAN}) to translate daytime vehicle images into nighttime style images for training detectors under poor lighting.
Paper \cite{Li2021C3GANConsistencyGAN} developed a Class-conditioned Consistency GAN for anomaly detection that conditions generation on class labels and incorporates consistency regularization. It applies anomaly detection in traffic data, enhancing the quality of synthetic anomaly samples for classifier training. By adopting this framework, the authors supersede vanilla GAN architectures which often run into model collapse.

% The most recent and efficient of all networks includes the Diffusion model (\cite{Song2020DDIM}). 
In recent years, diffusion-based deep generative models have shown remarkable success in capturing complex high-dimensional, multi-modal distributions \cite{Ho2020DenoisingDD,Song2021aScoreBasedGM,Kawar2022DenoisingDM,Xiao2022TacklingTI,Dhariwal2021DiffusionMW,Song2019GenerativeMW}. Several recent works employed diffusion models for generative modeling in Intelligent Transportation Systems. For instance, paper \cite{You2024Crossfusor} introduced Attention-based Video Diffusion to synthesize accident and non-accident datasets. It further applies the generated data to train an equivariant traffic accident anticipation model.

\begin{table*}[ht]
\centering
\caption{Comparative analysis of related works and the proposed model.}
\label{tab:related_works_comparison}
\begin{tabular}{@{}lcccc@{}}
\toprule
\textbf{Work} & \textbf{Core Methodology} & \textbf{\begin{tabular}[c]{@{}c@{}}Generative\\ Modeling\end{tabular}} & \textbf{Diffusion} & \textbf{\begin{tabular}[c]{@{}c@{}}High\\ Accuracy\end{tabular}} \\ \midrule
Jiang et al. \cite{small_parallel_residual_cnn_traffic_congestion} & Convolutional Neural Network (CNN) & $\times$ & $\times$ & \RIGHTcircle \\ \\
Nusari et al. \cite{review_yolo_cnn_real_time_accident_detection} & YOLO + CNN & $\times$ & $\times$ & \RIGHTcircle \\ \\
Jartarghar et al.  \cite{vision_transformer_multi_phase_accident_detection} & YOLO + ByteTrack + Vision Transformer (ViT) & $\times$ & $\times$ & \RIGHTcircle \\ \\
Xi et al. \cite{Zhang2023GANEnhancedTrafficAccidentDetection} & Generative Adversarial Network (GAN) + CNN/ViT & \checkmark & $\times$ & \RIGHTcircle \\ \\
Wang et al. \cite{Li2021C3GANConsistencyGAN} & Class-conditioned Consistency GAN (C3-GAN) & \checkmark & $\times$ & \RIGHTcircle \\ \\
% You et al. \cite{You2024Crossfusor} & Attention-based Video Diffusion Model & \checkmark & \checkmark & \RIGHTcircle \\ 
\midrule
\textbf{Proposed Model} & \textbf{Guided Diffusion Model + ExceptionNet} & \textbf{\checkmark} & \textbf{\checkmark} & \textbf{\checkmark} \\ \bottomrule
\end{tabular}
\end{table*}

Paper \cite{Lei2024ConditionalDiffusionTraffic}) introduced a Conditional Diffusion Framework with Spatio-Temporal Estimator (CDSTE) to predict dynamic traffic states and evolution with uncertainty modeling and (\cite{Zhang2024STCDM}) utilizes observable traffic data as conditional information in the reverse process of diffusion for traffic flow prediction. Paper \cite{You2024Crossfusor} introduced a transformer based diffusion model with a combination of enhanced cross-attention to model detailed car-following dynamics. This causes improved accuracy for long-term vehicle trajectory forecasting in autonomous driving contexts. Paper \cite{Wang2025KnowledgeGuidedDiffusion} introduced knowledge-guided conditioning to diffusion for generating realistic, controllable traffic data useful for simulation and anomaly detection studies. Paper \cite{Lu2025TrafficDataImputation} introduced a novel implicit-explicit conditional diffusion model designed for imputing missing entries in incomplete traffic datasets. They do this by using the features in a dual-stream architecture that separately captures the temporal characteristics and frequency domain patterns. 

We hypothesize that due to their high generalizability and scope of representation learning, diffusion models supersede previous discriminative models in  classification tasks. Paper \cite{Hoogeboom2021ArgmaxDM} introduced extensions of diffusion models for categorical data, and \cite{Austin2021DiffusionMF} have proposed diffusion models for discrete data as a generalization of the multinomial diffusion models, which could provide an alternative way of performing classification with diffusion-based models. To the best of our knowledge, this is the first paper which employs diffusion models for accident detection in order to achieve superior performance compared to the existing models. Table \ref{tab:related_works_comparison} presents a comparative analysis of related works and the proposed model.
% They also employ a cross-attention mechanism which merges these complementary features along with spatial adjacency information to improve the conditional guidance. Their method leverages the diffusion model’s inherent denoising capabilities and its non-recursive nature to minimize error accumulation.

% \begin{figure*}
    
% \end{figure*}

% \begin{figure}[b!]  % Force bottom placement
%     \centering
%     \includegraphics[width=0.7\textwidth,trim={0.5cm 3cm 0.5cm 3cm},clip]{lite.pdf}
%     \caption{Architecture of the proposed Diffusion model}
%     \label{fig:model_architecture}
% \end{figure}

\section{Proposed Methodology}\label{method}
\begin{figure*}
    \centering
    \includegraphics[width=1\linewidth]{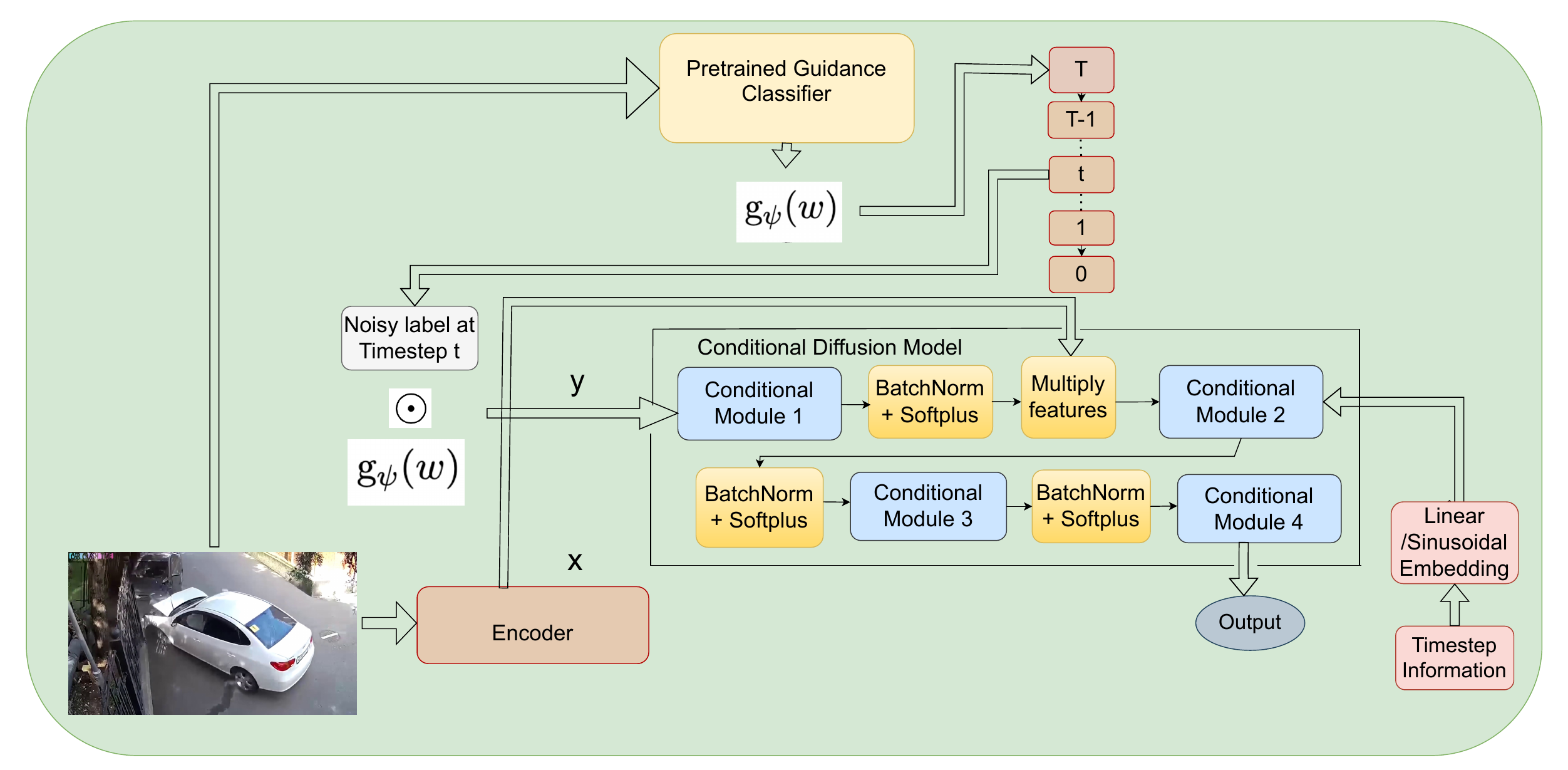}
    \caption{Architecture of the proposed Diffusion model}
    \label{fig:model_architecture}
\end{figure*}

Our proposed model architecture is illustrated in Figure~\ref{fig:model_architecture}, which provides a visual representation of the entire pipeline. The diagram shows how the guidance classifier's softmax outputs are integrated with the diffusion process, highlights the Conditional Module layers with their respective timestep embeddings, and demonstrates the crucial multiplication checkpoint between encoder features and processed labels. Figure~\ref{fig:model_architecture} also depicts the four Conditional Module blocks and their interconnections, offering a comprehensive overview of our hybrid classification approach.

The model is composed
of a guidance classifier, which is subjected to
classification via a diffusion process. The guidance classifier
consists of a modified architecture of a pre-trained ExceptionNet
model where the initial convolutional layer and the final fully connected layer are fine-tuned. The
fine-tuned layers of the model allow it to adapt to the
dataset.
% It is then followed by a Softmax output
layer in the architecture, which then trains the model.
Fine-tuning the model contributes to improved feature extraction
and generalization. These layers efficiently extract spatial
features using depthwise separable convolutions, as evidenced
by consistent performance across training and validation
metrics, hence improving classification performance (Section \ref{results}). The
pre-trained model weights of the guidance classifier are used in the diffusion classification.

Our model consists of multiple Conditional Modules. This component modulates and injects the linear projection of inputs using time-dependent embeddings, allowing the network to adapt its behavior dynamically throughout the diffusion process. Two types of embeddings are used:
\begin{itemize}
\item Learnable (Linear) Embeddings: Implemented via an embedding layer, these embeddings are learned during training and provide flexibility in representing the diffusion timestep.
\item Fixed (Sinusoidal) Embeddings: Fixed sinusoidal embeddings, inspired by the positional encodings in Transformer architectures, enable the model to more effectively distinguish between different noise levels during the denoising process.
% Cite the trasnformer paper
\end{itemize}
\subsection{Classification Through Diffusion Models}

Training diffusion models involves optimizing the evidence lower bound (ELBO), which can be decomposed into a series of KL divergences between the forward and reverse processes. This is done at each timestep and hence allows the model to train Classification through diffusion models involving conditioning on covariates $w$ and the predictions from the guidance classifier. We extend the concept of Denoising Diffusion Models\cite{ho2020ddpm} to Diffusion Models that have the ability to classify categories by integrating classification objectives into the diffusion process \cite{han2024card}. A model is trained such that the log-likelihood is maximized by optimizing the following ELBO:
\begin{align}
\log p_\theta (z_0 \mid w) &= \log \int p_\theta (z_{0:T} \mid w)\,dz_{1:T} \\
&\geq \mathbb{E}_{q(z_{1:T} \mid z_0, w)}\!\biggl[\log \frac{p_\theta (z_{0:T} \mid w)}{q(z_{1:T} \mid z_0, w)}\biggr]
\end{align}

where $q(z_{1:T} |z_0, w)$ is called the forward process or diffusion process in diffusion models. We denote $D_{KL}(q\parallel p)$ as the Kullback-Leibler (KL) divergence from distribution $p$ to distribution $q$. The objective can be rewritten as:

\begin{align}
L_{ELBO}(z_0, w) &:= L_0(z_0, w) + \sum_{t=2}^T L_{t-1}(z_0, w) + L_T(z_0, w) \label{eq:elbo} \\[6pt]
L_0(z_0, w) &:= E_q\left[-\log p_\theta(z_0 \mid z_1, w)\right] \\[6pt]
L_{t-1}(z_0, w) &:= E_q\left[D_{KL}\left(q(z_{t-1}\mid z_t, z_0, w) \right.\right. \nonumber\\
&\quad \left.\left.\|\,p_\theta(z_{t-1}\mid z_t, w)\right)\right] \\[6pt]
L_T(z_0, w) &:= E_q\left[D_{KL}\left(q(z_T \mid z_0, w)\,\|\,p(z_T \mid w)\right)\right]
\end{align}

We follow the convention to assume $L_T$ does not depend on any parameter and will be close to zero by carefully diffusing the observed response variable $z_0$ towards a pre-assumed distribution $p(z_T |w)$. The remaining terms will make the model $p_\theta (z_{t-1}|z_t, w)$ approximate the corresponding tractable ground-truth denoising transition step $q(z_{t-1}|z_t, z_0, w)$ for all timesteps. Unlike vanilla diffusion models, we assume the endpoint of our diffusion process to be:
\begin{equation}
p(z_T |w) = N (g_\psi (w), I)
\end{equation}

where $g_\psi (w)$ represents prior knowledge of the relation between $w$ and $z_0$, i.e. the pre-trained guidance classifier.

%  Describing: Algorithm Table:

% \subsection{Forward And Backward Process}

\begin{algorithm}[H]
\caption{Training (Classification)}\label{alg:classification}
\begin{algorithmic}[1]
\Require Pre-train $h_\phi(\mathbf{w})$ that predicts $\mathbb{E}(\mathbf{z} \mid \mathbf{w})$ with Cross Entropy loss
\Repeat
    \State Draw $\mathbf{z}_0 \sim q(\mathbf{z}_0 \mid \mathbf{w})$
    \State Draw $t \sim \text{Uniform}(\{1, \dots, T\})$
    \State Draw $\epsilon \sim \mathcal{N}(0, I)$
    \State Compute noise estimation loss:
    \[
    \mathcal{J}_\epsilon = \left\| \epsilon - \epsilon_\theta(\mathbf{w}, \sqrt{\bar{\rho}}\mathbf{z}_0 + \sqrt{1 - \bar{\rho}}g_\psi (\mathbf{w}), g_\psi (\mathbf{w}), t) \right\|^2
    \]
    \State Take numerical optimization step on: $\nabla_\theta \mathcal{J}_\epsilon$
\Until{Convergence}
\end{algorithmic}
\end{algorithm}

The objective detailed in ~\eqref{eq:elbo}) is optimized through an iterative training process, which is summarized in Algorithm~\ref{alg:classification} which outlines how the model learns to estimate the noise prediction $\epsilon_\theta$ by leveraging both the input image $\mathbf{u}$ and the guidance classifier's prediction $h_\phi(\mathbf{u})$.
Algorithm~\ref{alg:classification} details the training procedure for our proposed model. The process begins with a pre-training step for the guidance classifier $h_\phi(\mathbf{u})$, which is inferred from the trained the EfficientNet model. The core training loop iteratively draws a sample $\mathbf{z}_0$ from the conditional distribution, with the given image $\mathbf{w}$ as covariates, with a random timestep $t$. It also samples Gaussian noise $\epsilon$. The primary objective is to minimize the predicted noise, $\mathcal{J}_\epsilon$, which is the mean square error between the sampled noise $\epsilon$ and the noise predicted by the model $\epsilon_\theta$. 
% \subsection{Forward Diffusion Process}
In vanilla diffusion models, in the forward process, noise is incrementally added to the data over a series of time steps, effectively transforming structured data into noise. This process is typically modeled as a Markov chain, where each step adds Gaussian noise to the data:
\begin{equation}
q(z_t \mid z_{t-1}) = N\bigl(z_t; \sqrt{\delta_t}\,z_{t-1},\,(1 - \delta_t)I\bigr)
\end{equation}

The above can be extended to classification diffusion models. With a diffusion schedule \(\{\gamma_t\}_{t=1:T}\in(0,1)^T\), we specify the forward process conditional distributions similarly for all timesteps including \(t=1\):
\begin{align}
q(z_t \mid z_{t-1}, g_\psi (w)) &= \mathcal{N}\left(z_t;\, \sqrt{\delta_t}\,z_{t-1} + \left(1 - \sqrt{ \delta_t}\right)g_\psi (w),\right. \nonumber\\
&\quad \left.\delta_t I\right)
\end{align}
where,
\begin{align}
\gamma_t = 1 - \delta_t
\end{align}

Here, \(z_t\) represents the data at time step \(t\), \(\gamma_t\) is a variance schedule controlling the noise level at each step, and \(N\) denotes a Gaussian distribution. As \(t\) increases, \(z_t\) becomes progressively noisier, approaching a standard normal distribution as \(t\) approaches \(T\).

Let \(z_0\in\mathbb{R}^d\) be the original data. The forward diffusion process is defined as a Markov chain:
\begin{equation}
q(z_{1:T}\mid z_0) = \prod_{t=1}^T q(z_t \mid z_{t-1})
\end{equation}

Defining $\bar{\delta}_t = \prod_{i=1}^t \delta_i$, one can show by the properties of Gaussian distributions that the distribution of the noisy image at any timestep can be defined based on the original equation as follows
\begin{equation}
q(z_t \mid z_0) = N\bigl(z_t; \sqrt{\bar{\delta}_t}\,z_0,\,(1 - \bar{\delta}_t)I\bigr)
\end{equation}

% \subsection{Forward process sampling for classification diffusion model via cumulative alphas}

Similarly, for the conditional model, the forward process admits a closed-form sampling distribution at an arbitrary timestep $t$:
\begin{equation}
q(z_t|z_0, g_\psi (w)) = N (z_t; \sqrt{\bar{\delta}_t}z_0 + (1 - \sqrt{\bar{\delta}_t})g_\psi (w), (1 - \bar{\delta}_t)I)
\end{equation}

\

The forward process sampling distribution for any arbitrary timestep is obtained via the following process:

The expectation term is based on the forward process equation. Pandey et al.~\cite{pandey}. From earlier, we have that for all $t = 1, \ldots, T$,
\begin{align}
z_t &= \sqrt{1 - \gamma_t}z_{t-1} + (1 - \sqrt{1 - \gamma_t})g_\psi (w) + \sqrt{\gamma_t}\eta, \nonumber\\
&\quad \text{where } \eta \sim N (0, I)
\end{align}
Taking the expectations of both sides, we have

% \begin{align}
% E(z_t) &= \sqrt{1 - \gamma_t}E(z_{t-1}) + (1 - \sqrt{1 - \gamma_t})g_\psi (w) \\
% &= \sqrt{1 - \gamma_t}\left[\sqrt{1 - \gamma_{t-1}}E(z_{t-2}) + (1 - \sqrt{1 - \gamma_{t-1}})g_\psi (w)\right] \nonumber\\
% &\quad + (1 - \sqrt{1 - \gamma_t})g_\psi (w) \\
% &= \sqrt{(1 - \gamma_t)(1 - \gamma_{t-1})}E(z_{t-2}) + 
% (1 - \sqrt{(1 - \gamma_t)(1 - \gamma_{t-1})})g_\psi (w)
% \end{align}

\begin{align}
E(z_t) &= \sqrt{1 - \gamma_t} \, E(z_{t-1}) 
+ \left(1 - \sqrt{1 - \gamma_t}\right) g_\psi(w) \\
&= \sqrt{1 - \gamma_t} 
\left[ \sqrt{1 - \gamma_{t-1}} \, E(z_{t-2}) \right. \nonumber\\
&\quad \left. + \left(1 - \sqrt{1 - \gamma_{t-1}} \right) g_\psi(w) \right] \nonumber\\
&\quad + \left(1 - \sqrt{1 - \gamma_t} \right) g_\psi(w) \\
&= \sqrt{(1 - \gamma_t)(1 - \gamma_{t-1})} \, E(z_{t-2}) \nonumber\\
&\quad + \left(1 - \sqrt{(1 - \gamma_t)(1 - \gamma_{t-1})} \right) g_\psi(w)
\end{align}

Continuing this process recursively:
\begin{align}
&= \sqrt{\prod_{i=2}^t(1 - \gamma_i)}E(z_1) + \left(1 - \sqrt{\prod_{i=2}^t(1 - \gamma_i)}\right)g_\psi (w) \\
&= \sqrt{\prod_{i=2}^t(1 - \gamma_i)}\left[\sqrt{1 - \gamma_1}z_0 + (1 - \sqrt{1 - \gamma_1})g_\psi (w)\right] \nonumber\\
&\quad + \left(1 - \sqrt{\prod_{i=2}^t(1 - \gamma_i)}\right)g_\psi (w) \\
&= \sqrt{\prod_{i=1}^t(1 - \gamma_i)}z_0 + \left(1 - \sqrt{\prod_{i=1}^t(1 - \gamma_i)}\right)g_\psi (w) \\
&= \sqrt{\bar{\delta}_t}z_0 + (1 - \sqrt{\bar{\delta}_t})g_\psi (w)
\end{align}

% \subsection{Reverse Diffusion Process}
The mean term can be seen as an interpolation between true data $z_0$ and the predicted conditional expectation $g_\psi (w)$, which gradually shifts from the former to the latter throughout the forward process. \\

Further, in the reverse diffusion process, the goal is to learn the reverse of the diffusion process to generate new data samples from noise. The forward function process is as follows:
\begin{equation}
q(z_t|z_0) = N (z_t; \sqrt{\bar{\delta}_t}z_0, (1 - \bar{\delta}_t)I)
\end{equation}
The corresponding posterior $q(z_{t-1}|z_t, z_0)$ is also Gaussian:
\begin{equation}
q(z_{t-1}|z_t, z_0) = N (z_{t-1}; \tilde{\mu}_{\theta}(z_t, z_0), \tilde{\gamma}_t I)
\end{equation}
where
\begin{equation}
\tilde{\mu}_{t}(z_t, z_0) = \frac{\sqrt{\bar{\delta}_{t-1}}(1 - \delta_t)}{1 - \bar{\delta}_t}z_0 + \frac{\sqrt{\delta_t}(1 - \bar{\delta}_{t-1})}{1 - \bar{\delta}_t}z_t
\end{equation}
and
\begin{equation}
\tilde{\gamma}_t = \frac{1 - \bar{\delta}_{t-1}}{1 - \bar{\delta}_t}\gamma_t, 
\end{equation}

For the conditional distribution, the reverse process involves learning a parameterized model \(p_\theta(z_{t-1}\mid z_t)\) that approximates the reverse transitions. The parameterized model takes the following form:

\[
p_\theta(z_{t-1}\mid z_t)
= \mathcal{N}\bigl(z_{t-1};\,\mu_\theta(z_t, t),\,\Sigma_\theta(z_t, t)\bigr),
\]

where \(\mu_\theta\) and \(\Sigma_\theta\) are functions (often neural networks) that predict the mean and covariance of the reverse distribution. By sampling from this learned reverse process starting from pure noise \(z_T\sim\mathcal{N}(0,I)\), one can generate data resembling the original training data.

The reverse process is parameterized as:

\[
p_\theta(z_{0:T})
= p(z_T)\prod_{t=1}^T p_\theta(z_{t-1}\mid z_t).
\]

A common parameterization for the mean is given by:

\[
\mu_\theta(z_t, t)
= \frac{1}{\sqrt{\delta_t}}\,z_t
  - \frac{1 - \delta_t}{\sqrt{1 - \bar{\delta}_t}}\,\eta_\theta(z_t, t),
\]

where \(\eta_\theta(z_t, t)\) is a neural network predicting the noise component. 
Similarly for the classification diffusion models, the corresponding tractable forward process posterior is
\begin{align}
q(z_{t-1} | z_t, z_0, w) = q(z_{t-1} | z_t, z_0, g_\psi(w)) = \\
\mathcal{N}(z_{t-1}; \tilde{\mu}(z_t, z_0, g_\psi(w)), \tilde{\beta}_t I)
\end{align}
\begin{align}
\tilde{\mu} &:= \frac{\gamma_t\sqrt{\bar{\delta}_{t-1}}}{1 - \bar{\delta}_t}z_0 + \frac{(1 - \bar{\delta}_{t-1})}{1 - \bar{\delta}_t}z_t \nonumber\\
&\quad + \left(1 + \frac{(\sqrt{\bar{\delta}_t} - 1)(\sqrt{\bar{\delta}_{t-1}} - \sqrt{\bar{\delta}_t})}{1 - \bar{\delta}_t}\right)g_\psi (w) \\
\tilde{\gamma}_t &:= \frac{1 - \bar{\delta}_{t-1}}{1 - \bar{\delta}_t}\gamma_t
\end{align}

% \subsection{Derivation for Forward Process Posteriors} `In this section, 
We derive the mean and variance of the forward process posteriors $q(z_{t-1}|z_t, z_0, w)$, which are tractable when conditioned on $z_0$:
\begin{align}
&q(z_{t-1}|z_t, z_0, w) \propto q(z_t|z_{t-1}, g_\psi (w))q(z_{t-1}|z_0, g_\psi (w)) \nonumber\\
&\propto \exp\left( -\frac{1}{2}\left(\frac{(z_t - (1 - \sqrt{\delta_t})g_\psi (w) - \sqrt{\delta_t}z_{t-1})^2}{\gamma_t} \right.\right. \nonumber\\
&\quad\left.\left. + \frac{(z_{t-1} - \sqrt{\bar{\delta}_{t-1}}z_0 - (1 - \sqrt{\bar{\delta}_{t-1}})g_\psi (w))^2}{1 - \bar{\delta}_{t-1}}\right)\right)
\end{align}

This expands to:
\begin{align}
&\propto \exp\Bigg( -\frac{1}{2}\bigg(\frac{\delta_t z^2_{t-1} - 2\sqrt{\delta_t}z_t z_{t-1} - 2(1 - \sqrt{\delta_t})g_\psi (w)z_{t-1}}{\gamma_t} \nonumber\\
&\quad + \frac{z^2_{t-1} - 2\sqrt{\bar{\delta}_{t-1}}z_0 z_{t-1} - 2(1 - \sqrt{\bar{\delta}_{t-1}})g_\psi (w)z_{t-1}}{1 - \bar{\delta}_{t-1}}\bigg)\Bigg)
\end{align}

This simplifies to:
\begin{flalign*}
&\exp\Biggl(-\frac{1}{2}\Biggl(\Bigl(\frac{\delta_t}{\gamma_t} + \frac{1}{1-\bar{\delta}_{t-1}}\Bigr)z_{t-1}^2 - 2\Bigl(\frac{\sqrt{\bar{\delta}_{t-1}}}{1-\bar{\delta}_{t-1}}z_0 + \frac{\sqrt{\delta_t}}{\gamma_t}z_t \\
&\quad + \Bigl(\frac{\sqrt{\delta_t}(\sqrt{\delta_t}-1)}{\gamma_t} + \frac{1-\sqrt{\bar{\delta}_{t-1}}}{1-\bar{\delta}_{t-1}}\Bigr)g_{\psi}(w)\Bigr)z_{t-1}\Biggr)\Biggr) &&
\end{flalign*}

where 
\begin{align}
\frac{\delta_t(1-\bar{\delta}_{t-1}) + \gamma_t}{\gamma_t(1-\bar{\delta}_{t-1})} = \frac{1-\bar{\delta}_t}{\gamma_t(1-\bar{\delta}_{t-1})}
\end{align}
and we have the posterior variance
\begin{align}
\tilde{\gamma}_t = \frac{1}{1-\frac{\bar{\delta}_t}{1-\bar{\delta}_{t-1}}}\gamma_t
\end{align}
Meanwhile, the following coefficients of the terms in
the posterior mean by dividing each coefficient by
$\frac{1-\bar{\delta}_t}{1-\bar{\delta}_{t-1}}$:
\begin{align}
\lambda_0 &= \frac{\sqrt{\bar{\delta}_{t-1}}}{1-\bar{\delta}_{t-1}} \cdot \frac{1-\bar{\delta}_{t-1}}{1-\bar{\delta}_t} = \frac{\sqrt{\bar{\delta}_{t-1}}}{1-\bar{\delta}_t}\gamma_t\\
\lambda_1 &= \frac{\sqrt{\delta_t}}{\gamma_t} \cdot \frac{1-\bar{\delta}_{t-1}}{1-\bar{\delta}_t} = \frac{1-\bar{\delta}_{t-1}}{1-\bar{\delta}_t}\sqrt{\delta_t}
\end{align}

and
\begin{align}
\lambda_2 &= \left(\frac{\sqrt{\delta_t}(\sqrt{\delta_t}-1)}{\gamma_t} + \frac{1-\sqrt{\bar{\delta}_{t-1}}}{1-\bar{\delta}_{t-1}}\right) \cdot \frac{1-\bar{\delta}_{t-1}}{1-\bar{\delta}_t}\\
&= \frac{\delta_t - \bar{\delta}_t - \sqrt{\delta_t}(1-\bar{\delta}_{t-1}) + \gamma_t - \gamma_t\sqrt{\bar{\delta}_t}}{1-\bar{\delta}_t}\\
&= 1 + \frac{(\sqrt{\bar{\delta_t}}-1)(\sqrt{{\delta}_t}-\sqrt{\bar{\delta}_{t-1}})}{1-\bar{\delta}_t}
\end{align}
which together give us the posterior mean
\begin{align}
\tilde{\mu}(z_t, z_0, g_{\psi}(w)) = \lambda_0 z_0 + \lambda_1 z_t + \lambda_2 g_{\psi}(w)
\end{align}

\section{Performance Evaluation}\label{results}
\subsection{Dataset and Implementation}
We have used a publicly available dataset \cite{Accidentdset} to evaluate the proposed model. The images in the dataset are of size 128X128. The dataset comprises two classes, namely Accident and non-accident. We use 
training, validation, and testing splits of 80\%, 10\%, and 10\%, respectively. The proposed model achieved an accuracy of 97.32\% on the test set.
% The dataset has an effective dimensionality of \textbf{49152} with an input channel size of \textbf{3} and no additional padding. 
% The proposed model's architecture is a simple one that incorporates a feature dimension and a hidden dimension, both set to \textbf{4096}, ensuring sufficient capacity for high-dimensional data representation.

We have used data augmentation techniques like centre crop, random horizontal flip, and random rotation in the diffusion model training to reduce the problem of overfitting.
For the hyperparameters of the proposed model, we set the number of timesteps as 10, and we used a cosine noise schedule with its parameters - β1 and βT set to $10^{-4}$ and 0.02, respectively. A linear timestep embedding was also used.
% \section*{Hyperparameter Tuning of Our Proposed Model}
% A cosine schedule is used for the diffusion process and a linear embedding scheme is used for timesteps in the diffusion model.
For optimization, the Adam optimizer was used with a learning rate of 0.001 and a momentum parameter of 0.9. The weight decay parameter was set to 0, and the learning rate scheduling was enabled to adaptively adjust the learning rate throughout training. To maintain stable gradients, a gradient clipping threshold of 1.0 was applied. The epsilon parameter for numerical stability in the optimizer was set to 1e-8. These configurations ensured a robust and efficient training pipeline for the diffusion model.
% \section{Ablation Study}
We conducted a comprehensive evaluation of our model by comparing the training and testing performance across several metrics, including accuracy, precision, recall, F1-score, Mean Squared Error (MSE), and Cross Entropy. 
% For the training dataset, these metrics are plotted over epochs, while the testing dataset performance is summarized in tabular form in Table~\ref{tab:maintable}.

\begin{table*}[htbp]
\centering
\begin{tabular}{|l|c|c|c|c|c|c|}
\hline
\textbf{Model} & \textbf{Accuracy (\%)} & \textbf{Precision (\%)} & \textbf{Recall (\%)} & \textbf{F1 Score (\%)} & \textbf{CrossEntropy} & \textbf{MSE} \\
\hline
GoogLeNet \cite{szegedy2015going} & 68.00 & 64.79 & 86.79 & 74.19 & 0.5999 & 0.2047 \\
ResNet18 \cite{he2016deep} & 72.00 & 76.60 & 67.92 & 72.00 & 0.6219 & 0.2162 \\
ResNet50 \cite{he2016deep} & 74.00 & 72.13 & 83.02 & 77.19 & 0.5179 & 0.1699 \\
\textbf{Proposed model} & \textbf{97.32} & \textbf{99.40} & \textbf{97.32} & \textbf{98.14} & \textbf{0.3525} & \textbf{0.0644} \\
\hline
\end{tabular}
\caption{Performance metrics on the test dataset of different models.}
\label{tab:maintable}
\end{table*}

\begin{figure}[h]
    \centering
    % Top row: NLL and MSE side by side (smaller)
    \begin{subfigure}[b]{0.5\textwidth}
        \centering
        \includegraphics[width=\linewidth]{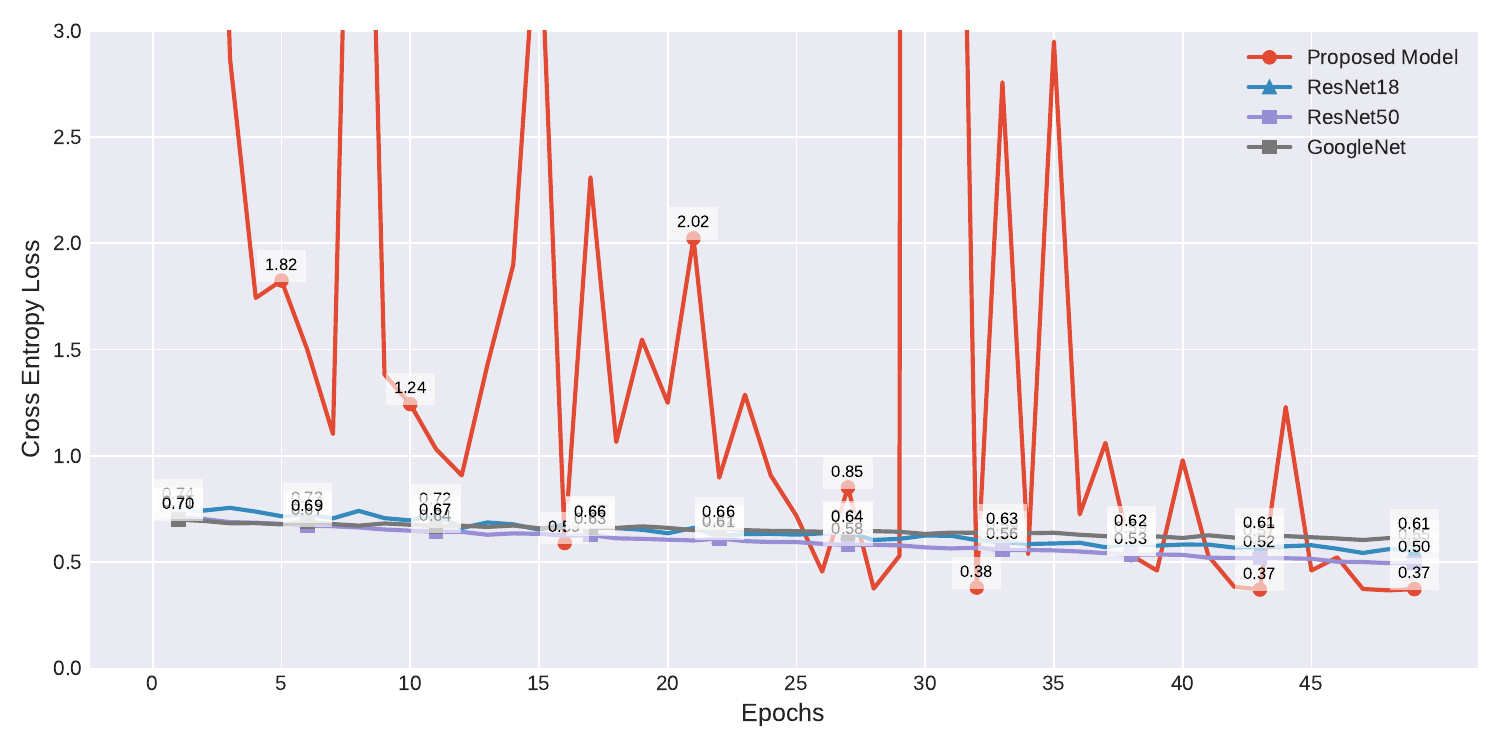}
        \caption{Cross Entropy}
        \label{fig:nll_2}
    \end{subfigure}
    \hfill
    \begin{subfigure}[b]{0.5\textwidth}
        \centering
        \includegraphics[width=\linewidth]{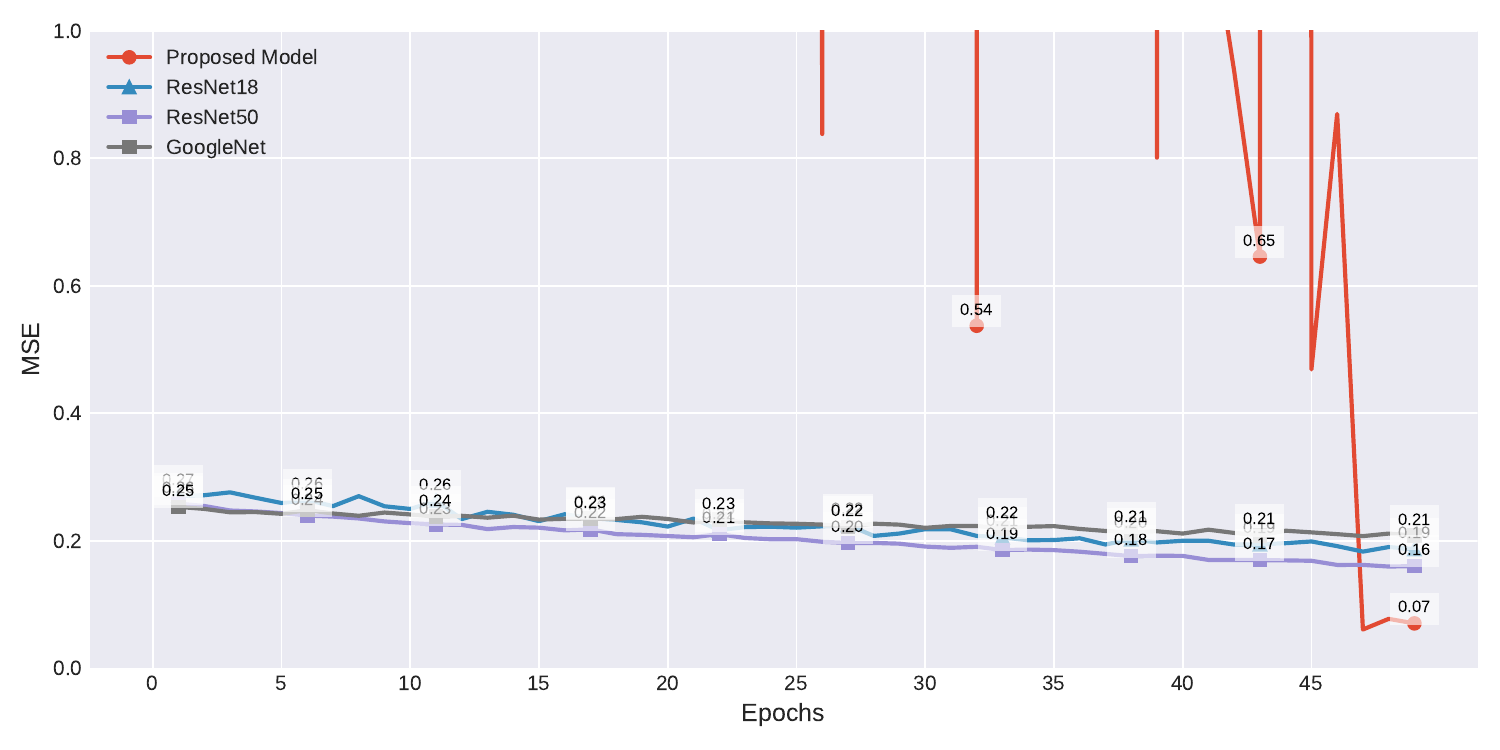}
        \caption{ MSE}
        \label{fig:mse}
    \end{subfigure}
    
    \vspace{0.1cm} % reduced space between rows
    
    % Bottom row: Train Accuracy (smaller)
    \begin{subfigure}[b]{0.5\textwidth}
        \centering
        \includegraphics[width=\linewidth]{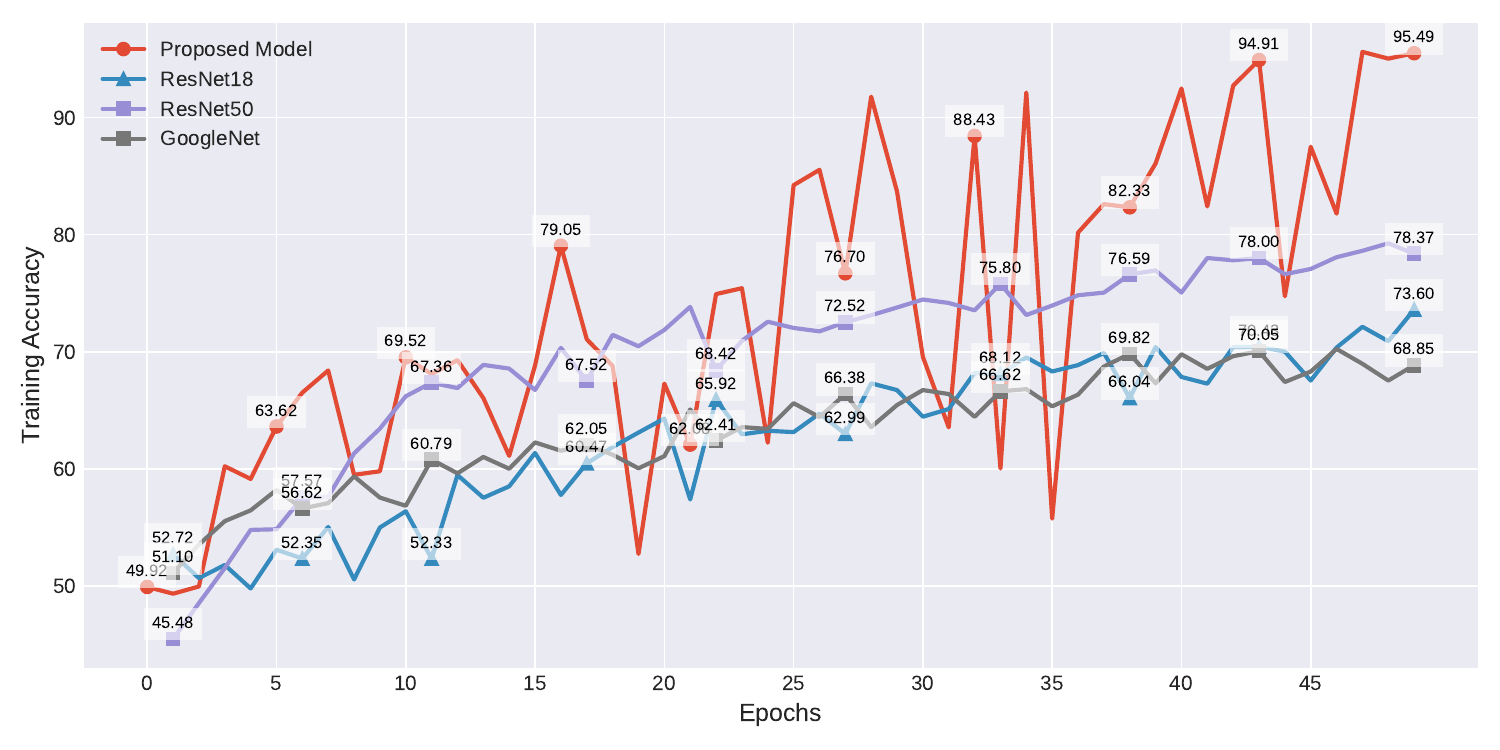}
        \caption{Accuracy}
        \label{fig:train_accuracy}
    \end{subfigure}
    \caption{Performance metrics across training epochs}
    \label{fig:all_metrics}
\end{figure}

\begin{table*}[]
\centering
\begin{tabular}{|l|l|l|l|l|l|l|}
\hline
\textbf{Model} & \textbf{Accuracy (\%)} & \textbf{Precision (\%)} & \textbf{Recall (\%)} & \textbf{F1 Score (\%)} & \textbf{CE} & \textbf{MSE} \\
\hline
Lin-Lin-Lin   & 91.07 & 99.40 & 91.10 & 94.70 & 0.387 & 0.113 \\
Lin-Lin-Sin   & 90.18 & 99.60 & 90.20 & 94.40 & 0.362 & 0.081 \\
Lin-Cos-Lin   & \textbf{97.32} & \textbf{99.40} & \textbf{97.30} & \textbf{98.10} & 0.353 & \textbf{0.064} \\
Lin-Cos-Sin   & 96.43 & 99.40 & 96.40 & 97.70 & 0.352 & 0.079 \\
Att-Lin-Lin   & 92.86 & 99.30 & 92.90 & 95.30 & 0.384 & 0.082 \\
Att-Lin-Sin   & 87.50 & 99.30 & 87.50 & 92.60 & 0.623 & 0.877 \\
Att-Cos-Lin   & 95.54 & 99.40 & 95.50 & 97.20 & \textbf{0.348} & 0.086 \\
Att-Cos-Sin   & 88.39 & 99.60 & 88.40 & 93.30 & 1.097 & 19.469 \\
\hline
\end{tabular}
\caption{Model performance comparison across different ablation configurations}
\label{tab:mainablationtable}
\end{table*}

\begin{figure*}[htbp]
    \centering

    % Row 1
    \begin{subfigure}{0.48\textwidth}
        \centering
        \includegraphics[width=\textwidth, trim=10 10 10 10, clip]{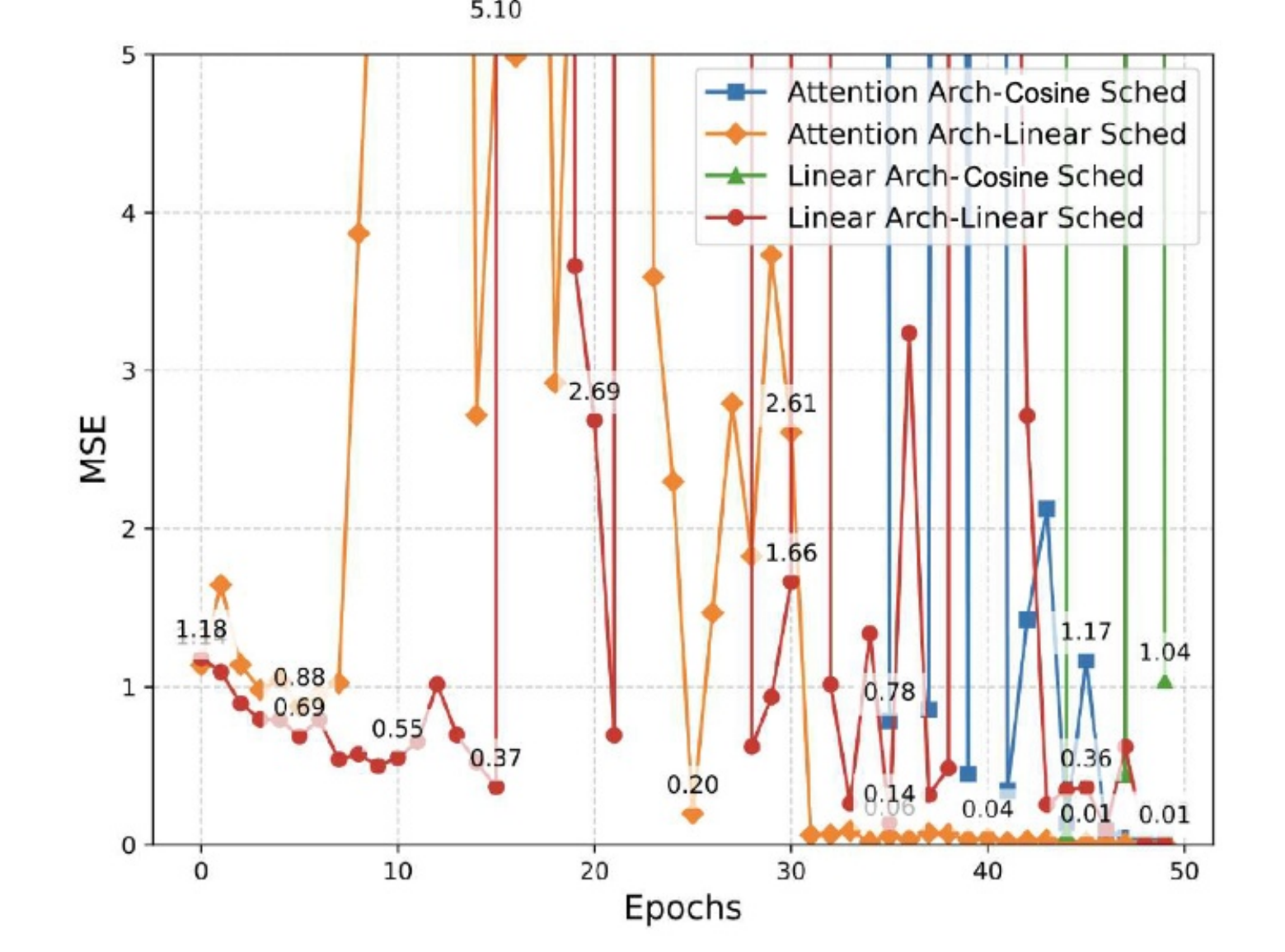}
        \caption{MSE-1}
        \label{fig:mse1}
    \end{subfigure}
    \hfill
    \begin{subfigure}{0.48\textwidth}
        \centering
        \includegraphics[width=\textwidth, trim=10 200 10 200, clip]{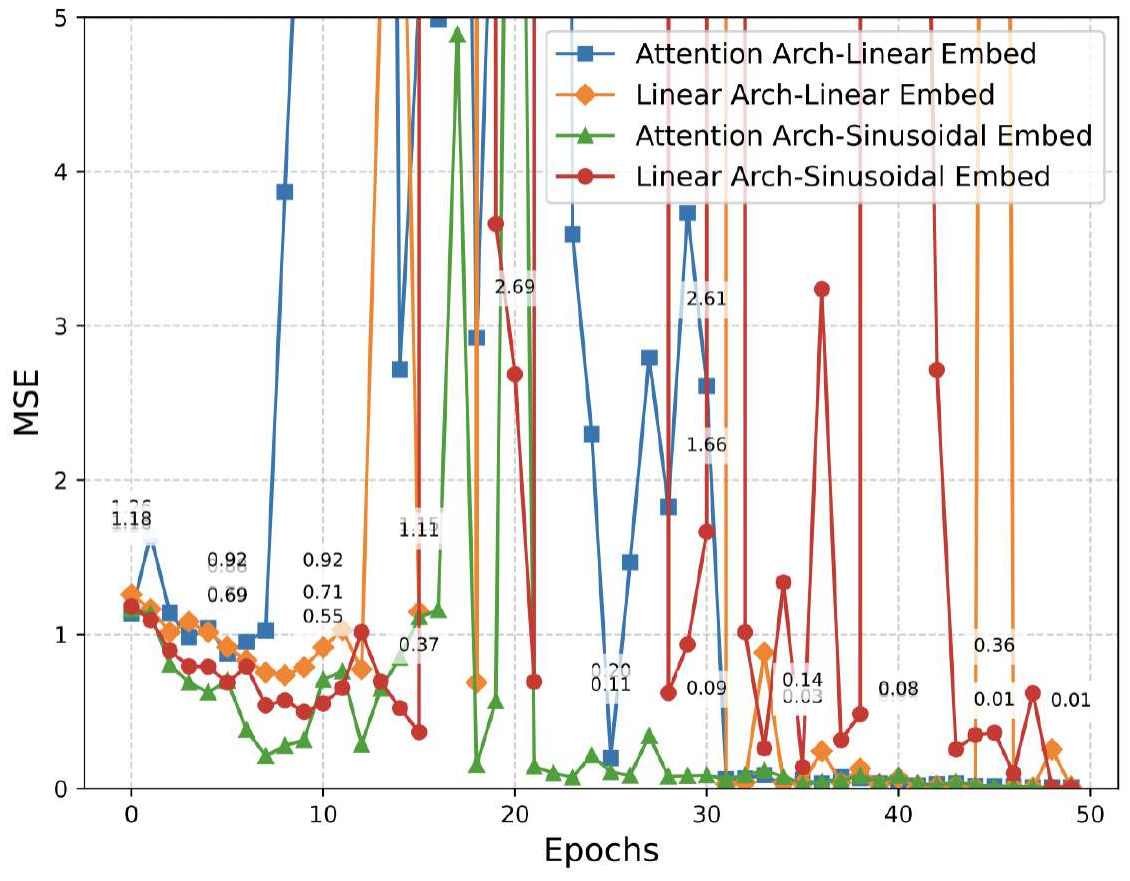}
        \caption{MSE-2}
        \label{fig:mse2}
    \end{subfigure}

    \vspace{0.5cm}

    % Row 2
    \begin{subfigure}{0.48\textwidth}
        \centering
        \includegraphics[width=\textwidth, trim=10 200 10 200, clip]{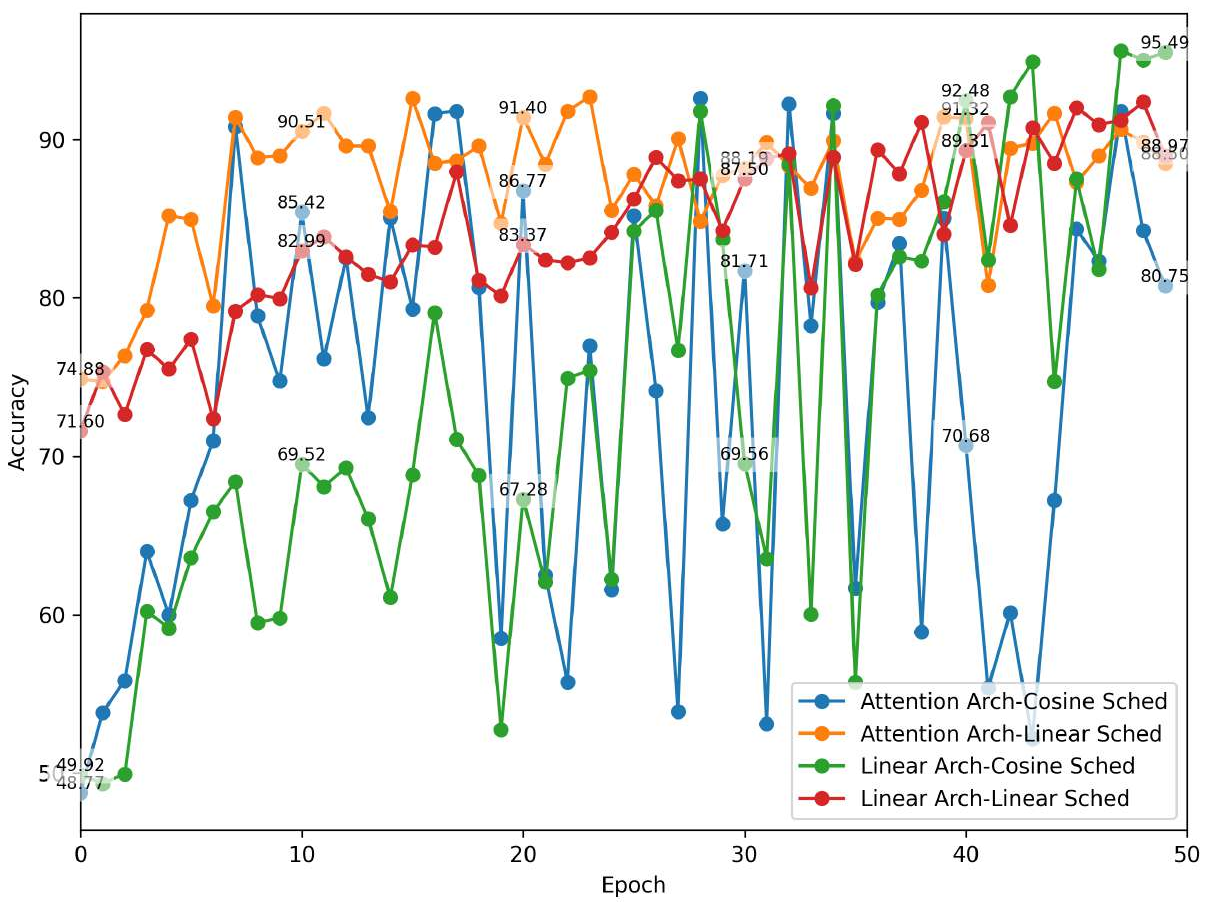}
        \caption{Accuracy-1}
        \label{fig:ac1}
    \end{subfigure}
    \hfill
    \begin{subfigure}{0.48\textwidth}
        \centering
        \includegraphics[width=\textwidth, trim=10 200 0 200, clip]{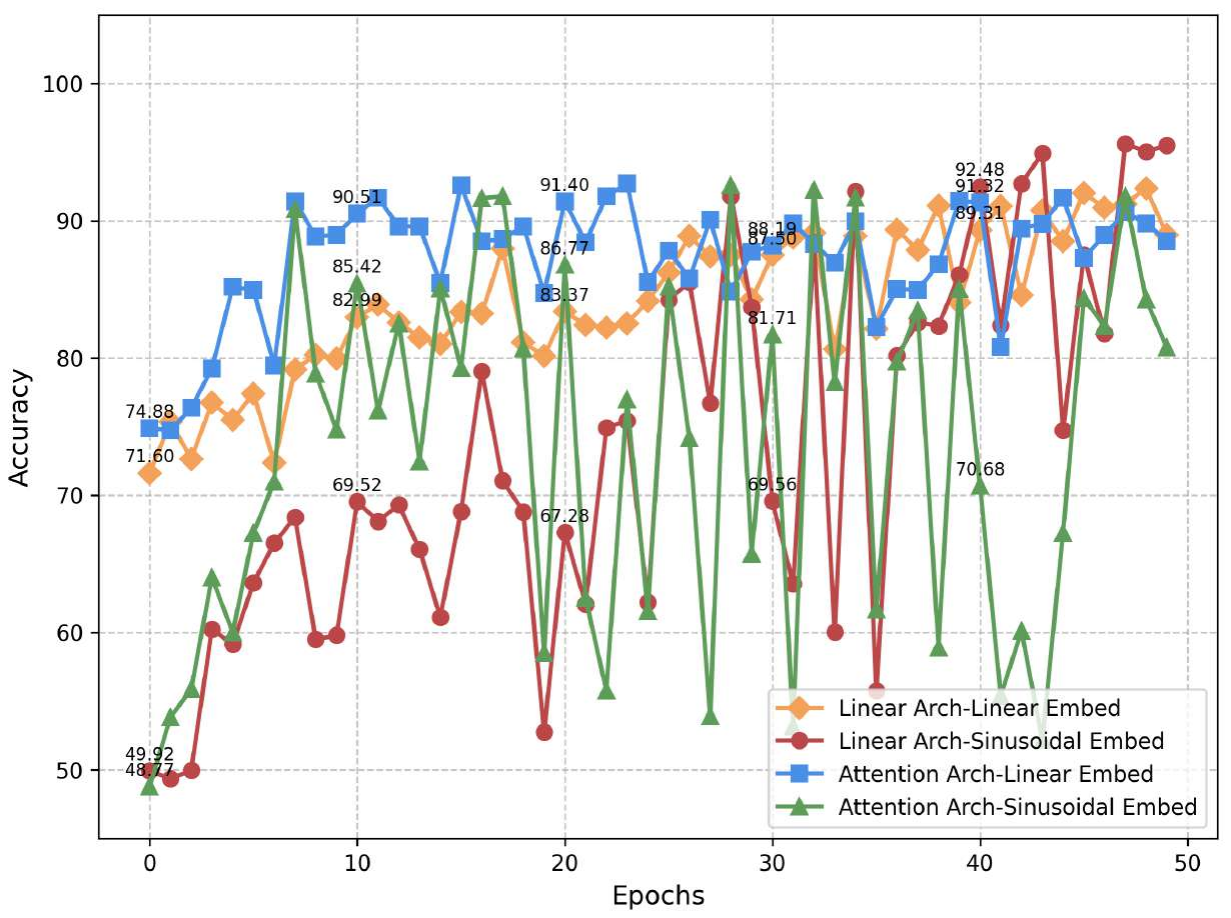}
        \caption{Accuracy-2}
        \label{fig:ac2}
    \end{subfigure}

    \vspace{0.5cm}

    % Row 3
    \begin{subfigure}{0.48\textwidth}
        \centering
        \includegraphics[width=\textwidth, trim=10 10 10 10, clip]{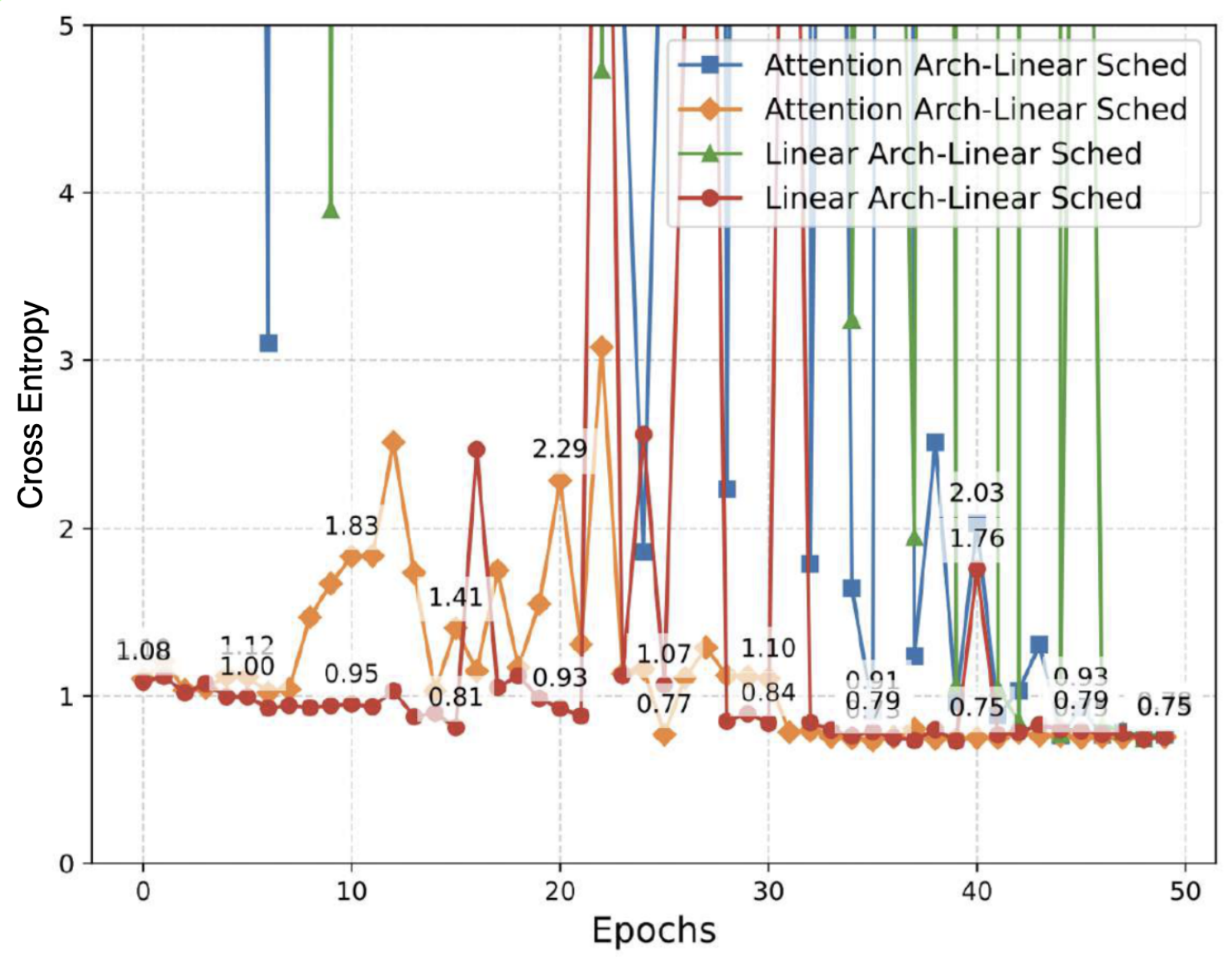}
        \caption{Cross Entropy-1}
        \label{fig:ce1}
    \end{subfigure}
    \hfill
    \begin{subfigure}{0.48\textwidth}
        \centering
        \includegraphics[width=\textwidth, trim=10 200 0 200, clip]{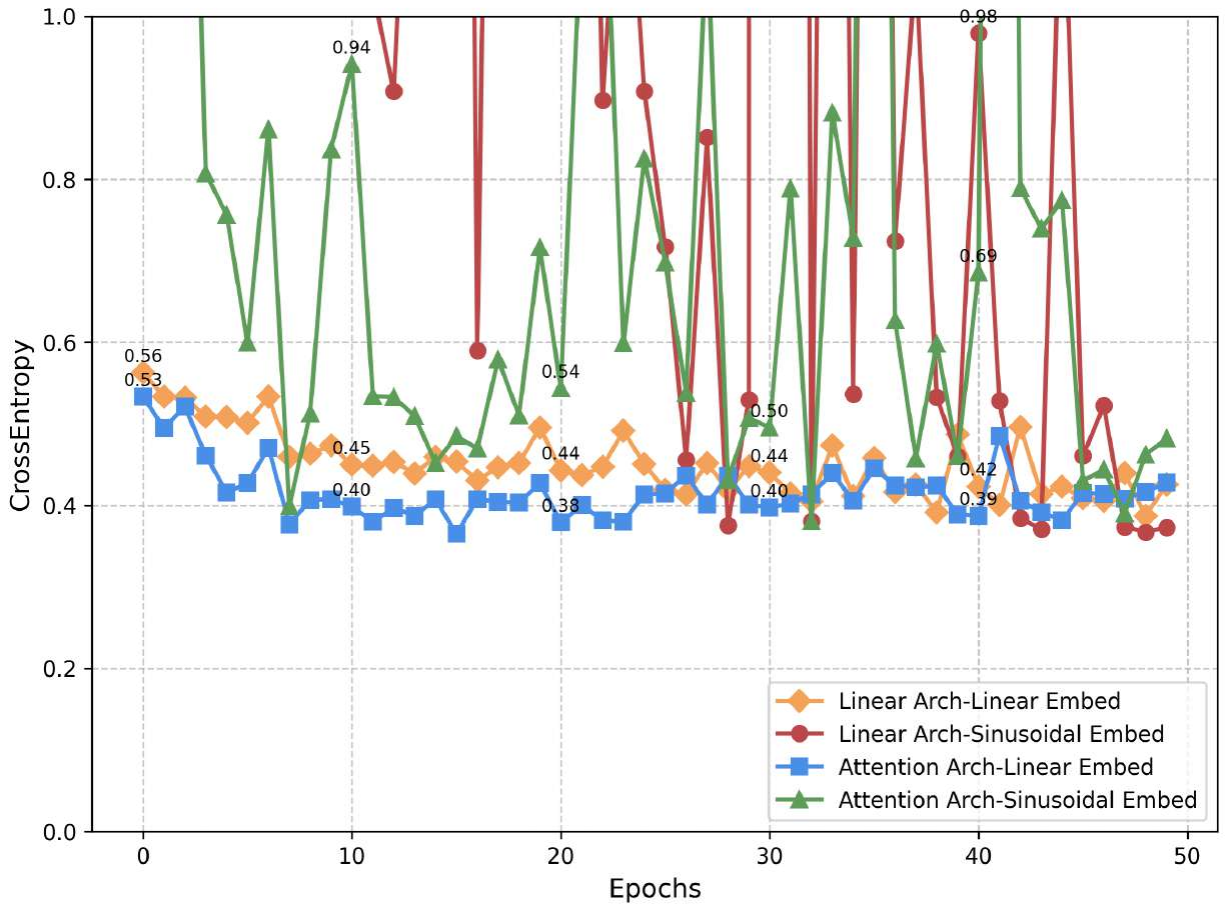}
        \caption{Cross Entropy-2}
        \label{fig:ce2}
    \end{subfigure}

    \caption{Comparison of model metrics of ablation studies}
    \label{fig:mainablationfigure}
\end{figure*}

The performance of the proposed model and four baselines are shown in Table~\ref{tab:maintable}. 
The proposed classification diffusion model achieved the highest accuracy with an impressive value of 97.32\%. In contrast, the conventional architectures—GoogLeNet, ResNet18, and ResNet50—recorded accuracies of 93\%, 84\%, and 87\%, respectively. This substantial improvement highlights the exceptional overall performance of the proposed model in correctly classifying instances.

The proposed model demonstrated outstanding precision with a score of 99.40\%, which far exceeds the precision of the traditional models: GoogLeNet at 64.79\%, ResNet18 at 76.60\%, and ResNet50 at 72.13\%. This indicates that the proposed model is extremely effective in minimizing false positives.
In terms of recall, the proposed model again excelled, achieving a value of 97.32\%. The competing models showed lower recall values with GoogLeNet at 86.79\%, ResNet18 at 67.92\%, and ResNet50 at 83.02\%. 
% The higher recall signifies the proposed model's robust ability to identify the majority of true positive instances.
The F1-Score, which balances precision and recall, was highest for the proposed model at 98.14\%. In comparison, GoogLeNet, ResNet18, and ResNet50 achieved F1-Scores of 74.19\%, 72.00\%, and 77.19\% respectively. This further confirms the proposed model's superior performance in balancing both key aspects of classification accuracy.
The conventional models yielded the following CrossEntropy and Negative Log-Likelihood (NLL) values: GoogLeNet at 0.5999, ResNet18 at 0.6219, and ResNet50 at 0.5179. The proposed model, on the other hand, achieved a notably lower CrossEntropy loss and NLL of 0.3525. 
% This reduction in loss reflects the proposed model's enhanced capability to reduce the error between predicted and true labels.
Finally, the Mean Squared Error (MSE) for the proposed model was the lowest at 0.0644 compared to the traditional models: GoogLeNet at 0.2047, ResNet18 at 0.2162, and ResNet50 at 0.1699. 
% The low MSE further underscores the proposed model's superior performance in minimizing the discrepancy between the predictions and the actual labels.
Overall, the results clearly demonstrate that the proposed model not only surpasses the conventional architectures in all primary evaluation metrics but also offers a significant improvement in loss reduction, indicating its effectiveness and robustness in classification tasks.

\begin{table*}[t]
\centering
\begin{tabular}{|l|l|l|l|l|l|l|}
\hline
\textbf{Timesteps} & \textbf{Accuracy (\%)} & \textbf{Precision (\%)} & \textbf{Recall (\%)} & \textbf{F1 Score (\%)} & \textbf{CrossEntropy} & \textbf{MSE} \\
\hline
10 & 88.39 & 99.60 & 88.40 & 93.30 & 1.097 & 19.469 \\
20 & 95.54 & 99.30 & 95.50 & 97.10 & 0.378 & 0.062 \\
30 & 95.54 & 99.30 & 95.50 & 96.70 & 0.347 & 0.051 \\
\hline
\end{tabular}
\caption{Performance metrics for different time steps}
\label{tab:timestepstable}

\end{table*}

% \section{Describing the Plots}\label{conclusion}
The training performance of the proposed model and the baseline models was evaluated over 50 epochs, as depicted in Figure \ref{fig:all_metrics}.
% \begin{enumerate}[leftmargin=*,noitemsep]
    % \item 
    As shown in Figure~\ref{fig:nll_2}, 
    the proposed model and ResNet50 exhibit lower and more stable cross-entropy loss values throughout training, indicating more consistent learning. ResNet18 and GoogLeNet, on the other hand, show larger fluctuations, suggesting instability during training. By the final epochs, all models converge to relatively low loss values, but the proposed model and ResNet50 maintain lower peaks and fewer fluctuations, underscoring their robust performance.
     Figure~\ref{fig:mse} shows that the proposed model converges to very low MSE values; although it exhibits occasional high-error spikes, ResNet50 maintains the lowest MSE values overall, demonstrating consistent performance with minimal fluctuations. GoogLeNet follows a similar pattern but displays slightly higher variability at certain epochs. ResNet18, while competitive, exhibits slightly higher fluctuations compared to ResNet50.
     Figure~\ref{fig:train_accuracy} illustrates the training accuracy progression for each model. Our model consistently achieved high accuracy, surpassing 95\% after approximately 30 epochs and outperforming the other models after just 20 epochs. It stabilizes near 100\% towards the end of training. ResNet18 followed a similar trend but displayed slightly more variability early on.
% \end{enumerate}

% \section*{Experiments}

\subsection{Ablation Studies}
In our experiments, we systematically varied key parameters of the proposed model, specifically:
% \begin{itemize}
%     \item Timestep embeddings used, namely \emph{sinusoidal} or \emph{linear}.
%     \item The network architecture used.
%     \item The scheduling method used to encode timestep information.
%     \item Timestep Count
% \end{itemize}

\subsubsection{Timestep Embedding}
The choice between sinusoidal and linear embeddings affects how the model encodes temporal information. We used linear and sinusoidal embeddings.
\subsubsection{Network Architecture}
The network architecture determines how the encoder encodes the image data. We varied between sequential linear layers and the multi-head attention framework.

\begin{figure}[]
    \centering

    % Cross Entropy Plot
    \begin{subfigure}{\columnwidth}
        \centering
        \includegraphics[width=\columnwidth, trim=70 100 50 100, clip]{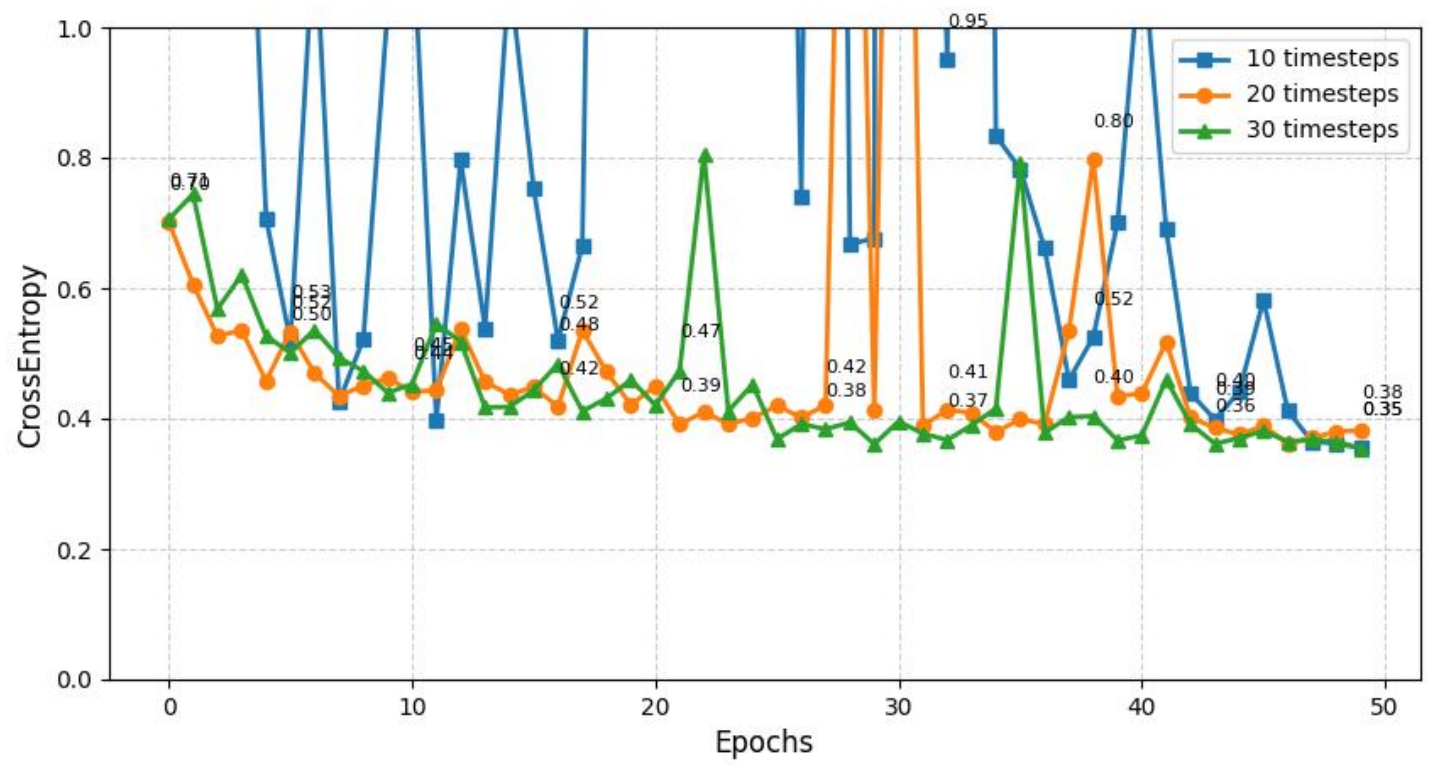}
        \caption{Cross Entropy over Timesteps}
        \label{fig:cross-entropy}
    \end{subfigure}

    \vspace{0.3cm} % Adjust spacing between subfigures

    % Accuracy Plot
    \begin{subfigure}{\columnwidth}
        \centering
        \includegraphics[width=\columnwidth, trim=25 100 0 100, clip]{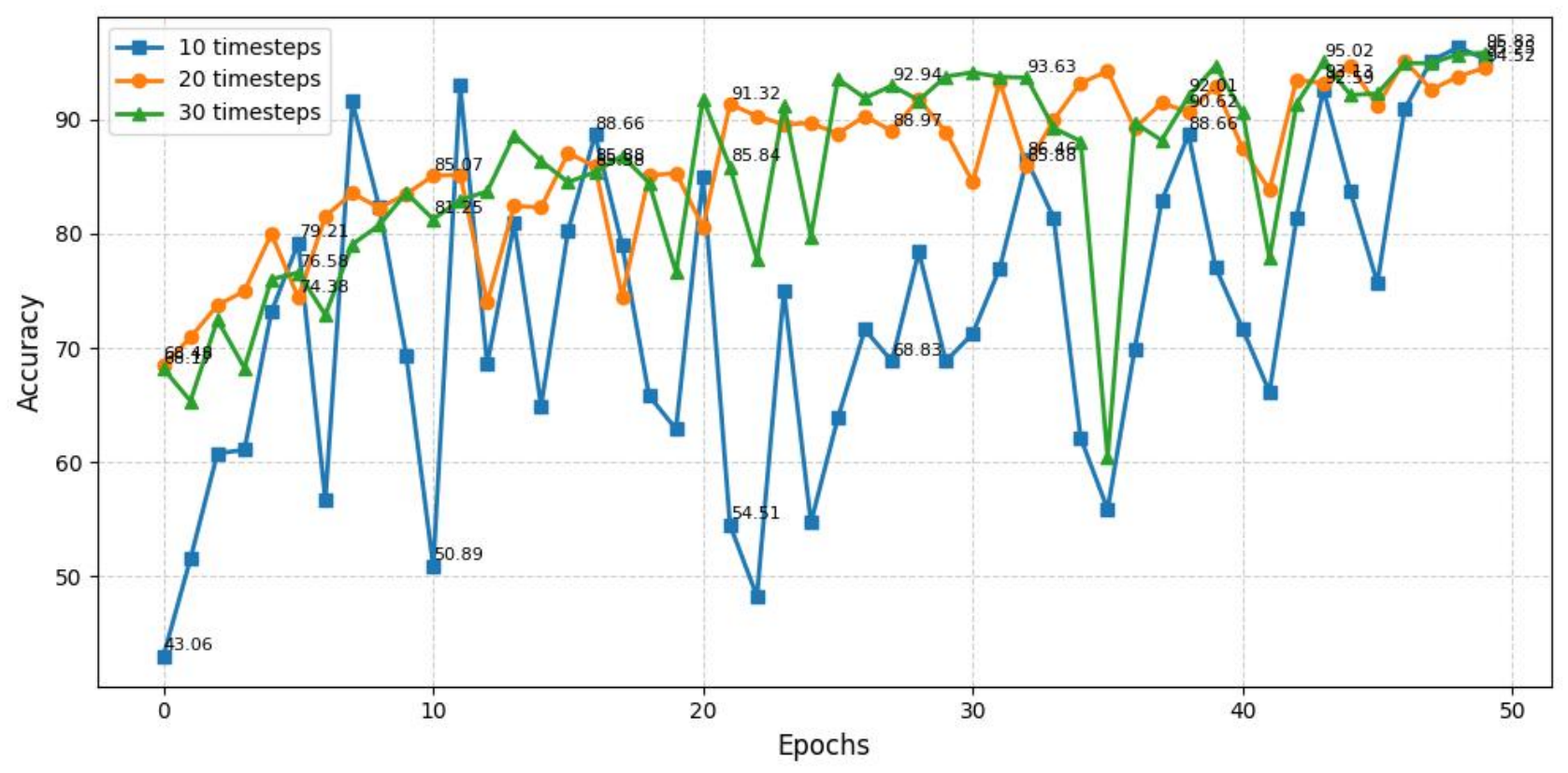}
        \caption{Accuracy over Timesteps}
        \label{fig:accuracy_timesteps}
    \end{subfigure}

    \vspace{0.3cm} % Adjust spacing between subfigures

    % MSE Plot
    \begin{subfigure}{\columnwidth}
        \centering
        \includegraphics[width=\columnwidth, trim=70 100 50 100, clip]{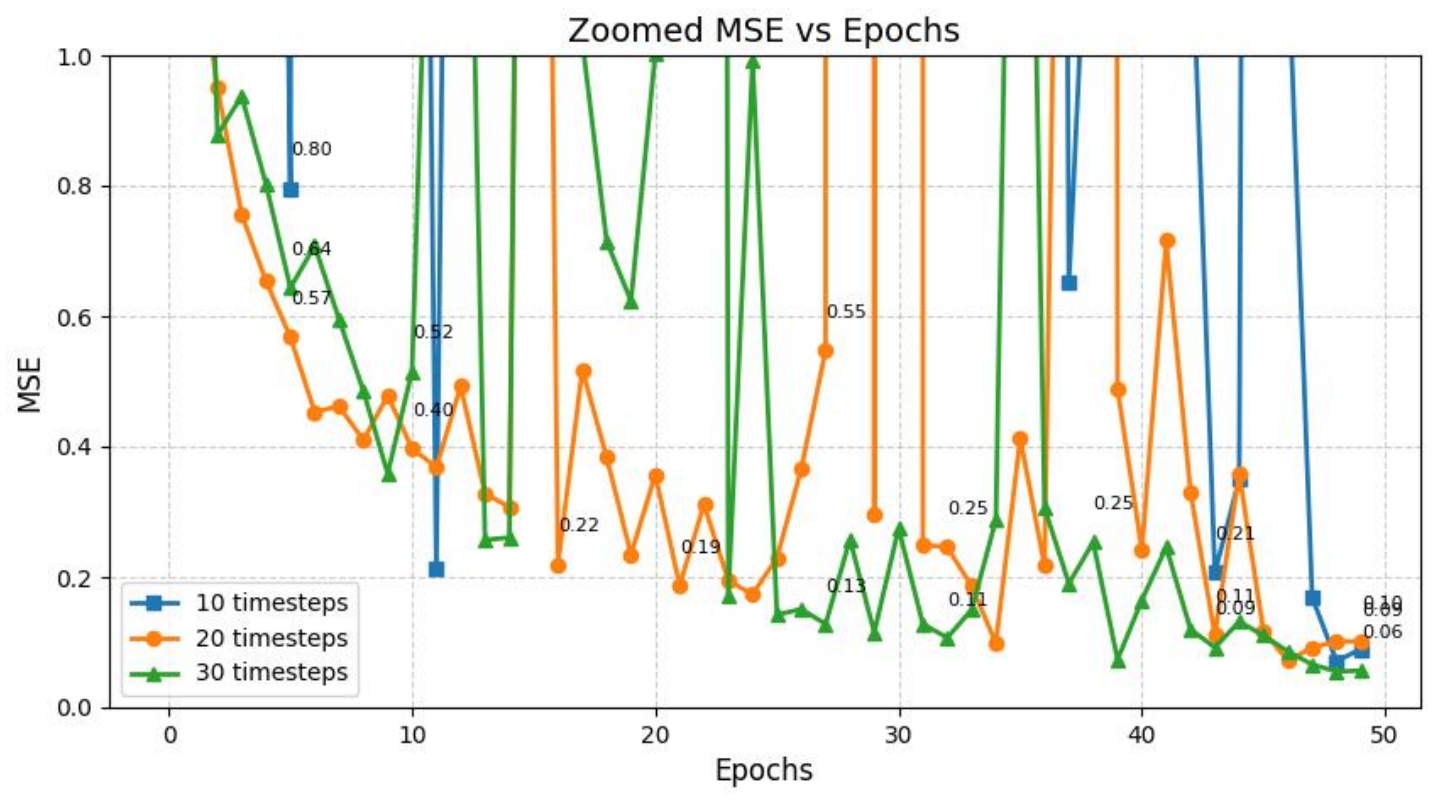}
        \caption{Mean Squared Error (MSE) over Timesteps}
        \label{fig:mse_timesteps}
    \end{subfigure}

    \caption{Performance Metrics over Timesteps}
    \label{fig:metrics_over_timesteps}
\end{figure}

\subsubsection{Scheduling Method}
The scheduling method refers to the beta schedule, which determines the amount of noise added across time steps in the diffusion process. The cosine-beta schedule strikes an optimal balance by avoiding insufficient noise addition in the early stages and excessive noise in the final stages while allowing a substantial increase in noise during the middle phase. This ensures that samples at any timestep are equally valuable to the training process, while a linear schedule adds noise at a steady, uniform rate throughout the process.

\subsubsection{Timestep Count}
The number of time steps used in the training process.

% All experiments were conducted using a fixed specification of $t = 10$ timesteps.
For the architecture, we adopted multi-head attention blocks. 
% In the \textbf{attention-based encoder}—designed for image data—
Herein, the model first divides the input image into patches via a convolutional layer, with the kernel size set equal to the desired patch size. These patches are then flattened and treated as tokens. A multi-head attention block processes these tokens, enabling the model to capture global dependencies and contextual relationships across the image. The attention outputs are aggregated using a token combiner (which includes layer normalization and a linear transformation) to generate a robust feature representation.

% It is also important to note that our paper includes comprehensive plots. 

Table \ref{tab:mainablationtable} presents accuracy, cross-entropy loss, MSE, precision, recall and F1-score across all the model variations based on the ablation parameters. These variations are as follows:
%\begin{enumerate}[noitemsep,leftmargin=*]
    varying time embeddings (linear and sinusoidal) with linear architecture and a linear time schedule;
varying time embeddings (linear and sinusoidal) with linear architecture and a cosine time schedule;
varying time embeddings (linear and sinusoidal) with attention-based architecture and a linear time schedule, and 
varying time embeddings (linear and sinusoidal) with attention-based architecture and a cosine time schedule.
  % variations in the training accuracy, Cross Entropy loss, and mean squared error (MSE) plots
% tables evaluating metrics on the Testing Dataset, namely Test Accuracy, Cross Entropy Loss, MSE, Precision, Recall and F1-Score across all model variations in Table~\ref{tab:model_performance_comparison}.
% \end{enumerate}
The Lin-Cos-Lin configuration achieves the highest accuracy (97.32\%) and F1-score(98.10\%), making it the strongest performer overall. Other models like "Lin-Cos-Sin" and "Att-Cos-Lin" perform competitively in some metrics.
Figure~\ref{fig:mainablationfigure} plots the model variations in terms of training MSE, accuracy, and cross-entropy. In column 1 of Figure \ref{fig:mainablationfigure}(i.e., Figures \ref{fig:mse1}, \ref{fig:ac1}, and \ref{fig:ce1}), the type of scheduler is varied - linear time scheduler or cosine time scheduler. In column 2( Figures \ref{fig:mse2}, \ref{fig:ac2}, and \ref{fig:ce2}), the variation is in the type of embedding- linear or sinusoidal time embedding.
% \end{itemize}
% \subsubsubsection{Column 1}

The MSE plot (Figures \ref{fig:mse1}, \ref{fig:mse2}) shows that the attention architecture with sinusoidal embedding converges faster. The accuracy (Figures \ref{fig:ac1}, \ref{fig:ac2}) and cross-entropy (Figures \ref{fig:ce1},\ref{fig:ce2}) show that linear embedding seems to learn faster across epochs and gives higher accuracy values early on compared to sinusoidal embedding. Across the graphs of all metrics, the sinusoidal embedding shows the highest variation across epochs for training accuracy. 
% \subsubsubsection{Column 2}
Figure \ref{fig:mse1} shows that the Mean Squared Error (MSE) converges the fastest for the attention architecture(linear schedule architecture and linear embedding). Both plots of the model employing linear schedule give higher accuracy values initially and show less variation overall, whereas cosine schedule employing plots show higher variation in accuracy.
% \section{Ablation Study on Timesteps}
% In this study, we evaluate the influence of different timesteps on the performance of our proposed diffusion model. 
We experimented with three configurations: 10, 20, and 30 timesteps, the results of which are listed in Table~\ref{tab:timestepstable}. The table shows that the test accuracy is notably higher at 20 and 30 timesteps (both 95.54\%). Further, the model with 30 timesteps yields lower cross-entropy (0.347) and MSE (0.051) and hence gives the best overall loss metrics.

The training metrics—accuracy, cross-entropy, and mean squared error (MSE) are illustrated in  Figure \ref{fig:metrics_over_timesteps}. The plots indicate that 20 and 30 timesteps generally yield better convergence in both accuracy and loss metrics, with 30 timesteps often giving the lowest MSE. The 10 timesteps plots show a higher variation in all metrics across the epochs.
These figures provide insights into how the model's learning behaviour varies with the number of time steps. 
% Additionally, Table~\ref{tab: timesteps} summarizes the testing performance across the different configurations.

\section{Conclusions and Future Work}\label{concl}
This paper proposed a novel diffusion-based classification model for vehicle accident detection that achieves 97.32\% accuracy, significantly outperforming traditional CNN architectures like GoogLeNet, ResNet18, and ResNet50. 
Our proposed model creates an efficient classification framework by leveraging a pre-trained neural network's outputs as diffusion means. The Conditional Modules in the proposed model modifies the linear projection of inputs using time-dependent embeddings. 
Our comprehensive ablation studies revealed optimal configurations with cosine scheduling and linear time embeddings. The model demonstrated exceptional precision (0.9940), recall (0.9732), and F1-score (0.9814) while maintaining lower loss values. Our findings suggest that diffusion models offer promising advancements for Intelligent Transportation Systems, with the potential to enhance real-time accident detection capabilities and, hence, improve road safety. As a part of future work, we plan to explore model compression through structural pruning, knowledge distillation, and quantization to enable edge AI deployment while retaining the high accuracy.

\bibliography{bibliography}
\bibliographystyle{IEEEtran}

\end{document}